\documentclass[sigconf,10pt]{acmart}

\renewcommand{\paragraph}[1]{\medskip\noindent\textbf{#1}}

\usepackage{xspace}

\definecolor{commentgreen}{rgb}{0, 0.5, 0}

\usepackage{float}
\usepackage[english]{babel}
\usepackage{subfig}
\usepackage[inline]{enumitem}
\usepackage{multirow}
\usepackage{comment}
\usepackage{booktabs}
\usepackage{caption}
\newcommand*\circled[1]{\tikz[baseline=(char.base)]{
		\node[shape=circle,draw,inner sep=0pt] (char) {#1};}}

\definecolor{diagramred}{RGB}{158, 11, 30}
\newcommand*\redcircled[1]{\tikz[baseline=(char.base)]{
		\node[shape=circle,draw=none,fill=diagramred,inner sep=0pt] (char) {\textcolor{white}{#1}};}}

\usepackage[scaled=0.85]{FiraMono}

\usepackage{minted}
\setminted{
	fontsize=\scriptsize,
	breaklines,
}

\usepackage{tcolorbox}
\tcbuselibrary{minted,skins}
\definecolor{codebg}{HTML}{F7F7F8}

\newenvironment{denseitemize}{
	\begin{itemize}[topsep=2pt, partopsep=0pt, leftmargin=1.5em]
		\setlength{\itemsep}{2pt}
		\setlength{\parskip}{0pt}
		\setlength{\parsep}{0pt}
	}{\end{itemize}}

\begin{document}

  \title[Cornserve: A Distributed Serving System for Any-to-Any Multimodal Models]{Cornserve: A Distributed Serving System for Any-to-Any Multimodal Models}

  \makeatletter
  \def\@titlefont{\LARGE\sffamily\bfseries}
  \def\@authorfont{\Large\normalfont}
  \def\@affiliationfont{\normalsize\normalfont}
  \makeatother
  \author{Jae-Won Chung$^{\text{1},\ast}$\enskip Jeff J. Ma$^{\text{1},\ast}$\enskip Jisang Ahn$^{\text{1}}$\enskip Yizhuo Liang$^{\text{2}}$\\Akshay Jajoo$^{\text{3}}$\enskip Myungjin Lee$^{\text{3}}$\enskip Mosharaf Chowdhury$^{\text{1}}$}
  \acmRefAuthors{Jae-Won Chung, Jeff J. Ma, Jisang Ahn, Yizhuo Liang, Akshay Jajoo, Myungjin Lee, and Mosharaf Chowdhury}
  \affiliation{\vspace{1mm} $^{\text{1}}$University of Michigan \enskip $^{\text{2}}$University of Southern California \enskip $^{\text{3}}$Cisco Research \country{}}
  \renewcommand{\shortauthors}{Chung and Ma et al.}

  \copyrightyear{2026}
  \acmYear{2026}
  \setcopyright{cc}
  \setcctype{by}
  \acmConference[CAIS '26]{ACM Conference on AI and Agentic Systems}{May 26--29, 2026}{San Jose, CA, USA}
  \acmBooktitle{ACM Conference on AI and Agentic Systems (CAIS '26), May 26--29, 2026, San Jose, CA, USA}
  \acmDOI{10.1145/3786335.3813209}
  \acmISBN{979-8-4007-2415-2/2026/05}



  \begin{abstract}
	{\renewcommand{\thefootnote}{\fnsymbol{footnote}}\footnotetext[1]{Equal contribution.}}
	Any-to-Any models are an emerging class of multimodal models that accept combinations of multimodal data (e.g., text, image, video, audio) as input and generate them as output.
Serving these models are challenging; different requests with different input and output modalities traverse different paths through the model computation graph, and each component of the model have different scaling characteristics.

We present Cornserve, a distributed serving system for generic Any-to-Any models.
Cornserve provides a flexible task abstraction for expressing Any-to-Any model computation graphs, enabling component disaggregation and independent scaling.
The distributed runtime dispatches compute to the data plane via an efficient record-and-replay execution model that keeps track of data dependencies, and forwards tensor data between components directly from the producer to the consumer.
Built on Kubernetes with approximately 23K new lines of Python, Cornserve supports diverse Any-to-Any models and delivers up to 3.81$\times$ higher throughput and 5.79$\times$ lower tail latency.
Cornserve is open-source,\footnote{\url{https://github.com/cornserve-ai/cornserve}} and the demo video is available on YouTube.\footnote{\url{https://www.youtube.com/watch?v=nb8R-vztLRg}}

  \end{abstract}

	\maketitle

	\section{Introduction}\label{sec:intro}

\looseness=-1
Our world is inherently multimodal, abound with data in the form of text, image, video, audio, and more.
Naturally, recent model developments generalize text-only Large Language Models (LLMs) and embrace \emph{multimodality}, leading to the emergence of a new class of models called \emph{Any-to-Any} models.
With over 11,000 variants on Hugging Face Hub as of March 2026~\cite{hf-any-to-any-models}, Any-to-Any models can
(1) understand multimodal inputs and/or
(2) generate multimodal outputs, with text being the most common modality.
For instance, Multimodal LLMs (MLLMs) like Qwen 2.5 VL~\cite{qwen2.5-vl-arxiv25} and InternVL 3~\cite{internvl3-arxiv25} can process multimodal inputs and generate text responses;
Qwen Image~\cite{qwen-image-arxiv25,qwen-image-2-blog} and GLM Image~\cite{glm-image-blog} produce images from text inputs processed by an LLM;
LTX-2~\cite{ltx2-arxiv26} generates audio and video;
DeepSeek Janus~\cite{janus-arxiv24,janus-pro-arxiv25} both understands and generates text and images;
and Qwen Omni~\cite{qwen2.5-omni-arxiv25,qwen3-omni-arxiv25} takes text, image, video, and audio inputs and generates text and audio.
Table~\ref{tab:a2a-models} summarizes modalities supported by recent Any-to-Any models.

\begin{table}[t]
  \footnotesize
  \centering
  \begin{tabular}{lll}
    \toprule
    \textbf{Model} & \textbf{Input} & \textbf{Output} \\
    \midrule
    Qwen 2.5 Omni~\cite{qwen2.5-omni-arxiv25}, Qwen 3 Omni~\cite{qwen3-omni-arxiv25} & T, I, V, A & T, A \\
    Qwen 2.5 VL~\cite{qwen2.5-vl-arxiv25}, InternVL 3~\cite{internvl3-arxiv25} & T, I, V & T \\
    DeepSeek Janus~\cite{janus-arxiv24,janus-pro-arxiv25} & T, I & T, I \\
    LTX-2~\cite{ltx2-arxiv26} & T, I & V, A \\
    Qwen Image~\cite{qwen-image-arxiv25}, GLM Image~\cite{glm-image-blog} & T & I \\
    \bottomrule
  \end{tabular}
  \vspace{0.75em}
  \caption{Modality breakdown of recent Any-to-Any multimodal models. The input and output modalities (\textbf{T}ext, \textbf{I}mage, \textbf{V}ideo, \textbf{A}udio) supported by a model can vary significantly.}\label{tab:a2a-models}
  \vspace{-2.0em}
\end{table}

\begin{figure}[t]
  \centering
  \subfloat[Multimodal input, text output (MLLM)]{
    \includegraphics[width=0.35\textwidth]{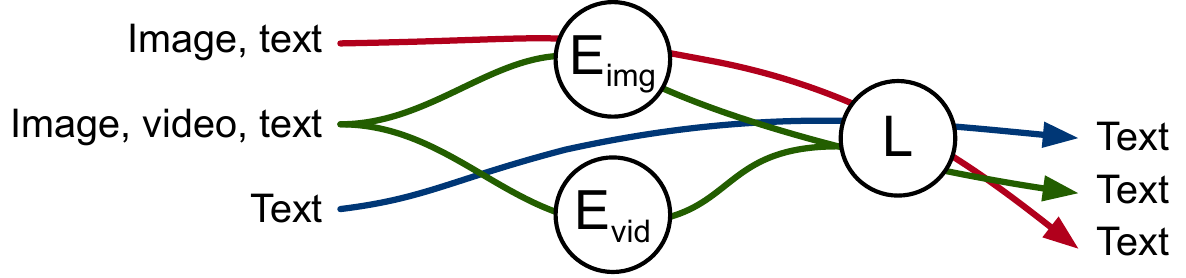}\label{fig:intro-a2a-mllm}
  }
  \vspace{-0.25em}
  \subfloat[Multimodal input, multimodal output]{
    \includegraphics[width=0.35\textwidth]{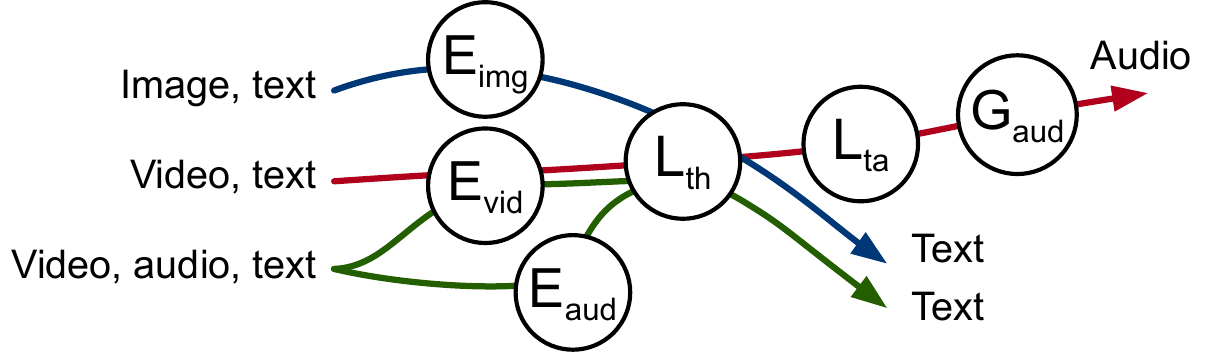}\label{fig:intro-a2a-omni}
  }
  \vspace{-0.5em}
  \caption{Computation graphs of (a) InternVL 3~\cite{internvl3-arxiv25}, a multimodal input model, and (b) Qwen Omni~\cite{qwen2.5-omni-arxiv25, qwen3-omni-arxiv25}, a multimodal input and output model. Different requests invoke different components and take different paths on the graph. $E$ stands for Encoder, $L$ for LLM, and $G$ for Generator. $L_{\text{th}}$ and $L_{\text{ta}}$ stand for thinker and talker LLMs, respectively.}
  \label{fig:intro-a2a-models}
  \vspace{-0.25em}
\end{figure}

Any-to-Any models can be viewed as a \emph{graph of heterogeneous components} that process different modalities, such as multimodal encoders, one or more LLMs, and multimodal generators (Figure~\ref{fig:intro-a2a-models}).
This introduces two types of heterogeneity to the serving system.
First is request type and computation path heterogeneity.
For instance, the Qwen Omni model in Figure~\ref{fig:intro-a2a-omni} consists of an image, video, and audio encoders that feed multimodal embeddings into the \emph{thinker} LLM for text generation.
If audio output is requested by the user, the text output and hidden states of the thinker LLM is passed to another autoregressive component called the talker that outputs audio tokens, which are subsequently decoded into waveforms by an audio generator (vocoder).
Thus, inference requests with different combinations of input and output modalities lead to different paths through the graph and invoke different model components, resulting in different request rates for each component.
This is exacerbated by the second type of heterogeneity: different components have different scaling characteristics and resource requirements.
For example, in Qwen 2.5 Omni, the text generation component achieves 4$\times$ higher throughput than the audio generator on the same A100 GPU.

Existing serving systems have offered at best \emph{point solutions} for Any-to-Any models.
Systems like vLLM~\cite{pagedattention-sosp23} have focused on text-only LLMs and MLLMs, whereas xDiT~\cite{xdit-arxiv24} has focused only on diffusion-based image or video generation, i.e., special cases of Any-to-Any models.
Similarly, techniques like Prefill--Decode (PD) disaggregation~\cite{distserve-osdi24,splitwise-isca24} and Encode--Prefill--Decode (EPD) disaggregation~\cite{epd-icml25} target specific model architectures and serving scenarios.

To this end, we present Cornserve, the first distributed serving system for generic Any-to-Any models, to the best of our knowledge.
Cornserve provides three key capabilities for efficiently serving Any-to-Any models:
\begin{denseitemize}
  \item A flexible \textbf{task abstraction} (unit tasks, composite tasks, and apps) that allows model developers to express arbitrary Any-to-Any model computations in plain Python.
  \item \textbf{Model fission} to disaggregate models at component boundaries into independently scalable components, each running on dedicated GPUs with specialized executors.
  \item A \textbf{distributed runtime} that supports models with arbitrary computation graphs through a record-and-replay execution mechanism, with efficient tensor data forwarding between components via shared memory and RDMA.
\end{denseitemize}
Additionally, when multiple models use equivalent components (e.g., a shared vision encoder), Cornserve automatically shares executor deployments to reduce GPU usage.

Cornserve is implemented on top of Kubernetes with approximately 23K lines of Python, with full support for OpenTelemetry for monitoring and tracing.
Cornserve supports a variety of recent Any-to-Any models including Qwen 2.5 VL~\cite{qwen2.5-vl-arxiv25}, InternVL 3~\cite{internvl3-arxiv25}, Qwen Image~\cite{qwen-image-arxiv25}, and Qwen 2.5/3 Omni~\cite{qwen3-omni-arxiv25}, and improves serving throughput by up to 3.81$\times$ and reduces tail latency by up to 5.79$\times$.

\section{System Design}
\label{sec:design}

\subsection{Architecture Overview}\label{sec:design-arch}

\begin{figure}
  \centering
  \includegraphics[width=0.45\textwidth]{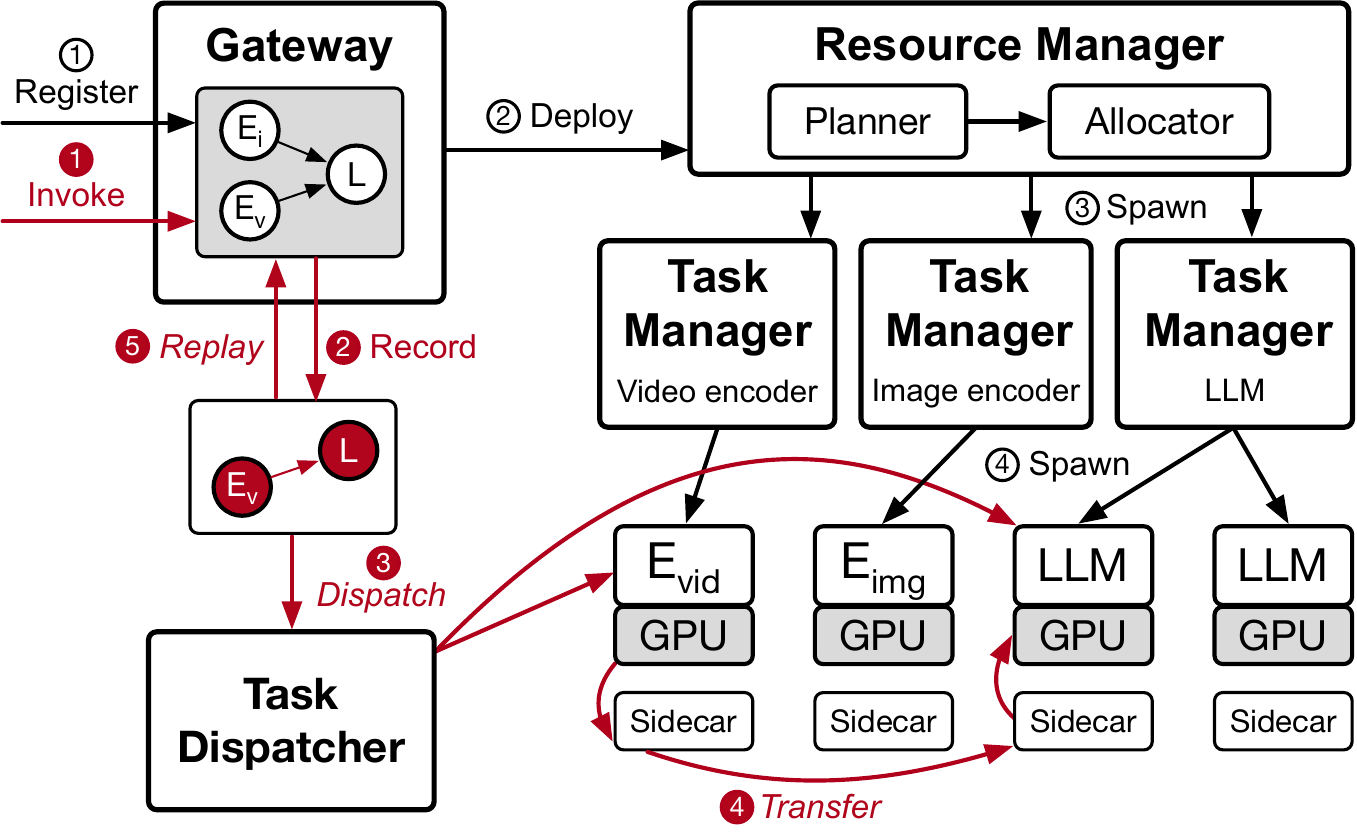}
  \vspace{-0.5em}
  \caption{
    Cornserve architecture and flows for deployment (black) and inference (red).}\label{fig:architecture}
\end{figure}

Figure~\ref{fig:architecture} illustrates Cornserve's overall architecture and how deployment and inference requests flow through the system.

\paragraph{Control Plane.}
The \emph{Gateway} serves as the entry point for \emph{App} registration and inference requests.
An App is a user-defined Python module that specifies one or more \emph{Tasks}, each of which specifies a model component and its dependencies on other components.
The \emph{Resource Manager} includes a \emph{planner}~\cite{cornfigurator-arxiv25} that finds an efficient deployment plan for tasks, and an \emph{allocator} that realizes the plan by spawning \emph{Task Managers}.
Task Managers (1) spawn \emph{Task Executors} (e.g., vLLM) on the allocated GPUs and (2) handles routing/load balancing.
The \emph{Task Dispatcher} routes component invocations from the Gateway to the appropriate Task Executors.

\paragraph{Data Plane.}
Each \emph{Task Executor} runs model components on GPUs.
Cornserve provides specialized executors for different component types: \emph{Eric} for multimodal encoders (image, video, audio), a forked \emph{vLLM}~\cite{pagedattention-sosp23} for LLM inference, and \emph{Geri} for multimodal generators (audio vocoder, image DiT).
The \emph{Sidecar}, which is a per-GPU daemon process, forwards tensor data between executors following data dependencies.

\paragraph{Deployment Lifecycle.}
An administrator \circled{1} registers an app by submitting a Python source file to the Gateway.
The Gateway validates the app, identifies its task definitions, and \circled{2} invokes the Resource Manager.
The Resource Manager \circled{3} spawns Task Managers, which in turn \circled{4} spawn Task Executor replicas on the allocated GPUs.

\paragraph{Request Lifecycle.}
When \redcircled{1} an inference request arrives at the Gateway, it \redcircled{2} identifies the specific components invoked by the request using a \emph{record-and-replay} approach (\S\ref{sec:design-task}) and \redcircled{3} forwards them to the Task Dispatcher, which dispatches these to the Task Executors.
Intermediate data \redcircled{4} flows through Sidecars, bypassing the control plane, and the final result is \redcircled{5} returned to the client.
All component invocations are dispatched simultaneously, and Task Executors wait for intermediate tensors to arrive through the Sidecar.

\subsection{Task Abstraction and Model Fission}
\label{sec:design-task}

Any-to-Any models exhibit request and computation path heterogeneity: depending on the request type, the model may invoke different components in different sequences.
For instance, in a multimodal LLM with an image encoder and an LLM, a request with both image and text invokes the encoder first and then the LLM, while a text-only request bypasses the encoder and directly invokes the LLM.
In the general case, this logic becomes arbitrarily complex with loops and branches that depend on the input, and the inference logic may include arbitrary processing before and after invoking a component (e.g., fitering returned text, modifying prompts).

\paragraph{Tasks and Apps.}
Cornserve introduces three levels of abstraction: \emph{unit tasks}, \emph{composite tasks}, and \emph{apps}.
A unit task, e.g., \texttt{LLMTask}, \texttt{EncoderTask}, \texttt{GeneratorTask}, represents an atomic model component, and is deployed as its own Task Manager.
Model developers compose unit tasks within a composite task class that implements a model's computations in plain Python inside the \texttt{invoke} method.
An \emph{app} is a single Python module that instantiates unit/composite tasks and defines an \texttt{async def serve} function as the entry point.

\paragraph{Model Fission.}
By expressing a model as a composite task of unit tasks, users can define \emph{fissions} at arbitrary component boundaries.
Listing~\ref{lst:composite-task} shows \texttt{MLLMTask}, a built-in composite task for multimodal LLMs.
When \texttt{encoder\_fission} is true, \texttt{post\_init} creates separate \texttt{EncoderTask} unit tasks that are deployed on dedicated GPUs, and the \texttt{invoke} method calls them to produce embeddings before invoking the LLM.
When false, the LLM handles encoding internally in a monolithic deployment.
Each unit task is associated with a \emph{task descriptor} that specifies how to deploy the Task Executor for that task.

\begin{listing}[t]
\begin{tcolorbox}[
  enhanced,
  colback=codebg,
  colframe=gray!40,
  boxrule=0.4pt,
  arc=1.5mm,
  left=2mm, right=2mm, top=0mm, bottom=1mm,
  title={\footnotesize\ttfamily cornserve\_tasklib/task/composite/llm.py},
  coltitle=black!90,
  fonttitle=\footnotesize\ttfamily,
  colbacktitle=gray!20,
  bottomtitle=0.4pt,
  titlerule=0pt,
]
\begin{minted}{python}
class MLLMTask(Task[ChatCompletion, Stream[str]]):
    model_id: str
    modalities: list[Modality]
    encoder_ids: set[str]
    encoder_fission: bool = True
  
    def post_init(self):
        if self.encoder_fission:
            self.encoders = {
                mod: EncoderTask(self.encoder_ids, modality)
                for modality in self.modalities
            }
        self.llm = LLMTask(
            self.model_id, recv_embeds=self.encoder_fission)
  
    def invoke(self, req: ChatCompletion) -> Stream[str]:
        if self.encoder_fission:
            for mes in extract_multimodal(req.messages):
                emb = self.encoders[mes.modality].invoke(mes)
                req.multimodal_embeddings.append(emb)
        return self.llm.invoke(req)
\end{minted}
\end{tcolorbox}
\caption{
  Simplified \texttt{MLLMTask} composite task.
  When \texttt{encoder\_fission} is true, \texttt{LLMTask} is instantiated with \texttt{recv\_embeds=True}, which is recognized by its task descriptor to disable the encoder and receive embeddings from the sidecar instead.
  Cornserve's record-and-replay captures unit task invocations and data dependencies automatically.}\label{lst:composite-task}
\end{listing}

\begin{listing}[t]
\begin{tcolorbox}[
  enhanced,
  colback=codebg,
  colframe=gray!40,
  boxrule=0.4pt,
  arc=1.5mm,
  left=2mm, right=2mm, top=0mm, bottom=1mm,
  title={\footnotesize\ttfamily examples/gemma\_arena.py},
  coltitle=black!90,
  fonttitle=\footnotesize\ttfamily,
  colbacktitle=gray!20,
  bottomtitle=0.4pt,
  titlerule=0pt,
]
\begin{minted}{python}
gemmas = ["google/gemma-3-4b-it", "google/gemma-3-12b-it"]
# All MLLMTasks share the same encoder deployment.
gemma_tasks = {
    model_id: MLLMTask(
        modalities=[Modality.IMAGE],
        model_id=model_id,
        encoder_ids=set(gemmas),
        encoder_fission=True,
    )
    for model_id in gemmas
}

class Config(AppConfig):
    tasks = {**gemma_tasks}

async def serve(req: ChatCompletion) -> AsyncIterator[dict]:
    streams = await asyncio.gather(
        *(task(req) for task in gemma_tasks.values())
    )
    async for model, chunk in merge_streams(gemmas, streams):
        yield {model: chunk}
\end{minted}
\end{tcolorbox}
\caption{Cornserve app serving multiple Gemma 3 models~\cite{gemma3-arxiv25}. The \texttt{serve} function defines the app's computation: it invokes multiple MLLM tasks concurrently, merges their streaming responses, and yields structured results, demonstrating how apps can express arbitrary logic in plain Python.}
\label{lst:gemma-arena}
\end{listing}

\paragraph{Record-and-Replay Execution.}
During runtime, for a given inference request, Cornserve must extract the subgraph of unit task invocations and dispatch them to the appropriate Task Executors.
The key challenge is that the invocation subgraph depends on the request content and may involve complex control flow, as illustrated in Listing~\ref{lst:composite-task}.

Cornserve addresses this with a two-phase approach.
In the \emph{record} phase, the composite task's \texttt{invoke} method is called with the request, but unit tasks return placeholder results while recording their invocations and inputs.
By matching placeholder object IDs across unit task inputs and outputs, Cornserve reconstructs the subgraph of the computation graph executed for the given request.
This phase is instantaneous, as no actual model computation is performed.
The recorded graph is then sent to the Task Dispatcher, which dispatches each unit task invocation to the appropriate executors and collects results.
In the \emph{replay} phase, the \texttt{invoke} method is called again, but now each unit task returns the real result from the dispatcher, producing the final response.

This record-and-replay approach allows Cornserve to support Any-to-Any models with most control flow constructs (e.g., loops, branches) while keeping the developer experience simple.
A limitation is that the composite task's control flow must be deterministic given the request, as \emph{replay} must follow the same path as the \emph{record}.
Dynamic data-dependent control flow (e.g., invoking different tasks based on an LLM's response) is handled at the app layer, where the \texttt{serve} function executes with real results and can have arbitrary logic.

\paragraph{Component Sharing.}
When multiple apps use equivalent unit tasks---same class and field values---they automatically share a single Task Manager and its executors.
Listing~\ref{lst:gemma-arena} shows a Cornserve app that serves multiple Gemma 3 models with a shared vision encoder.
Because all \texttt{MLLMTask} instances specify the same \texttt{encoder\_ids}, Cornserve deploys only one shared encoder Task Manager, reducing total GPU usage.

\paragraph{Data Forwarding.}
When model components are fissioned onto separate GPUs, intermediate data such as multimodal embeddings and hidden states must be transferred between executors via the Sidecar.
For intra-node transfers, Sidecars use Linux shared memory (\texttt{/dev/shm}) to minimize interference with NVLink/NVSwitch bandwidth, which is likely heavily utilized by executors running tensor or expert parallel inference~\cite{llumnix-osdi24}.
The sender writes intermediate data from GPU memory directly to shared memory and notifies the receiver of the data's address via gRPC.
Upon notification, the receiver reads the data directly from shared memory and copies it into its GPU memory.
For inter-node transfers, Sidecars use RDMA via the UCX communication library~\cite{ucx}.

	\section{Evaluation}\label{sec:evaluation}

Cornserve supports generic Any-to-Any models including MLLMs.
In this section, we highlight Qwen Omni models~\cite{qwen2.5-omni-arxiv25,qwen3-omni-arxiv25} as a generic and challenging model family to serve.

\paragraph{Setup.}
We conduct experiments on two 8$\times$ NVIDIA A100-80GB GPU nodes with NVSwitch and 400 Gbps cross-node bandwidth over RDMA.
We use ServeGen~\cite{servegen-nsdi26}, a real-world multimodal request dataset, and have 20\% requests generate audio outputs.
The baseline is a monolithic deployment where all components run within a single executor backed by Hugging Face Transformers~\cite{huggingface}, as vLLM~\cite{pagedattention-sosp23} does not support either Qwen Omni models fully.

\begin{figure}[t]
  \vspace{-1em}
  \begin{minipage}[b]{0.40\textwidth}
    \begin{figure}[H]
      \centering
      \includegraphics[width=0.9\linewidth]{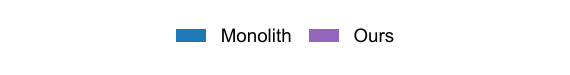}
      \vspace{-1.5em}
    \end{figure}
  \end{minipage}
  \setcounter{subfigure}{0}
  \centering
  \subfloat[Qwen 2.5 Omni Tput]{
    \includegraphics[width=0.32\linewidth]{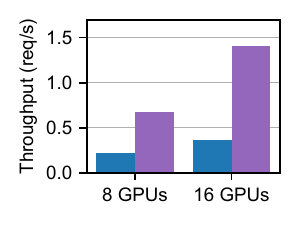}
    \label{fig:evaluation-qwen2.5-omni-tput}
  }
  \subfloat[Qwen 3 Omni Tput]{
    \includegraphics[width=0.32\linewidth]{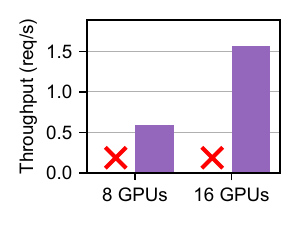}
    \label{fig:evaluation-qwen3-omni-tput}
  } 
  \subfloat[Qwen 2.5 Omni Lat.]{
    \includegraphics[width=0.30\linewidth]{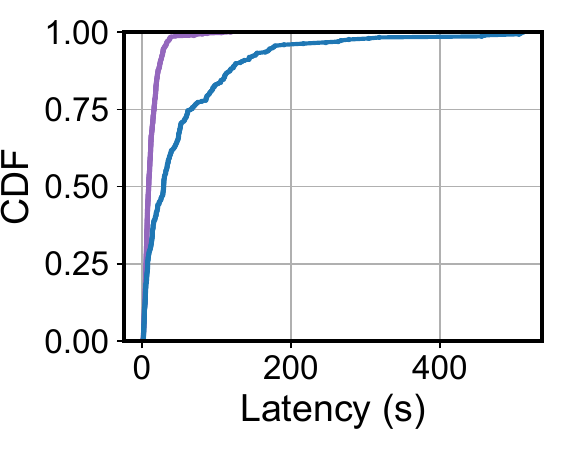}
    \label{fig:evaluation-qwen2.5-omni-latency}
  }
  \vspace{-0.5em}
  \caption{
    Monolith vs. Cornserve comparisons for Qwen 2.5 Omni 7B~\cite{qwen2.5-omni-arxiv25} throughput and latency CDF, and Qwen 3 Omni 30B~\cite{qwen3-omni-arxiv25} throughput.
    $\times$ indicate GPU out-of-memory.}\label{fig:evaluation-throughput}
\end{figure}

\paragraph{Throughput.}
Figures~\ref{fig:evaluation-qwen2.5-omni-tput} and~\ref{fig:evaluation-qwen3-omni-tput} compare serving Qwen 2.5 Omni and Qwen 3 Omni with monolith vs. Cornserve.
Cornserve improves the throughput of Qwen 2.5 Omni on 8-GPU and 16-GPU configurations by 3.09$\times$ and 3.81$\times$, respectively.
Due to the high computation scaling heterogeneity between the LLM and the audio generator, the planner~\cite{cornfigurator-arxiv25} replicates the audio generator independently (7 and 15 replicas on 8 and 16 GPUs, respectively) to balance throughput.

\begin{figure}[t]
  \vspace{-1em}
  \centering
  \subfloat[8 GPUs (1 node)]{
    \includegraphics[width=0.262\linewidth]{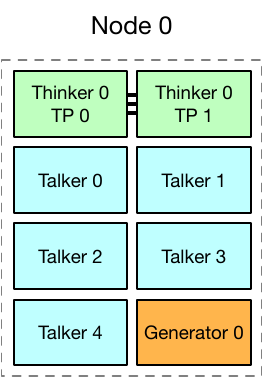}
    \label{fig:cell-8}
  }
  \hspace{1em}
  \subfloat[16 GPUs (2 nodes)]{
    \includegraphics[width=0.54\linewidth]{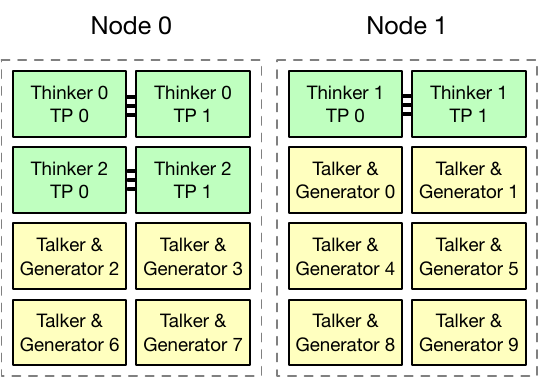}
    \label{fig:cell-16}
  }
  \vspace{-0.5em}
  \caption{Cornserve deployment configurations for Qwen 3 Omni on 8 and 16 GPUs. Each box represents a GPU. Model fission allows each component to scale independently: the thinker (LLM) uses tensor parallelism while talkers and generators are replicated to balance throughput.}
  \label{fig:cell-deployment}
\end{figure}

For Qwen 3 Omni, the monolithic deployment fails due to GPU out-of-memory errors, further highlighting the need for model fission.
Cornserve's planner~\cite{cornfigurator-arxiv25} disaggregates Qwen 3 Omni into separately scalable components, enabling efficient serving.
Figure~\ref{fig:cell-deployment} shows the resulting deployment configurations: on 8 GPUs, the thinker uses tensor parallelism across 2 GPUs while the talker is replicated across the remaining GPUs with one generator replica; on 16 GPUs, the thinker scales to 3 TP-2 replicas with 10 talker-generator replicas.
Scaling from 8 to 16 GPUs yields a 2.68$\times$ throughput improvement, exceeding the expected 2$\times$ from linear scaling as the additional GPUs allow better throughput balancing across components.

\paragraph{Latency.}
As shown in Figure~\ref{fig:evaluation-qwen2.5-omni-latency}, compared to monolithic deployment, Cornserve significantly reduces request latency for Qwen 2.5 Omni on 16 GPUs, with P50, P95, and P99 improvements of 3.24$\times$, 5.3$\times$, and 5.79$\times$, respectively.
These benefits stem from two sources.
First, in monolithic deployments, only one component of the model runs at a time, meaning whenever one component is running, all other components do not make progress.
Thus, removing this component interference improves latency by reducing blocking.
Second, disaggregating the audio generator allows the autoregressive audio generator to unlock continuous batching~\cite{orca-osdi22}, improving its throughput compared to the monolithic deployment; implementing talker continuous batching alongside the thinker LLM is technically challenging.

\paragraph{Sidecar Overhead.}
Sidecar transfers are not the main latency bottleneck as they only happen at disaggregated component boundaries.
Cross-node tensor forwarding via RDMA adds 5--10 ms for 8 MB, 8--20 ms for 16 MB, and 12--27 ms for 32 MB tensors at 5--15 transfers per second; intra-node forwarding via shared memory adds similar overhead.

	\section{Conclusion}
\label{sec:conclusion}

We present Cornserve, the first distributed serving system for generic Any-to-Any multimodal models.
Cornserve's task abstraction and model fission enable flexible and efficient deployment of complex Any-to-Any models.
We believe the strong inference serving platform provided by Cornserve will enable the development of more powerful and efficient Any-to-Any models, and we hope it will serve as a foundation for future research in this area.

	\begin{acks}
We thank the CAIS reviewers and members of SymbioticLab for their helpful discussions and feedback.
This work was supported in part by NSF grants CCF-2450085 and CNS-2106184, and by grants from Cisco, Ford, Mozilla Foundation, and Laude Institute.
Jae-Won Chung is additionally supported by the Kwanjeong Educational Foundation.
\end{acks}

  \label{EndOfPaper}

  \clearpage
	{
		\bibliographystyle{plain}
		\bibliography{ref}
	}

\end{document}